\documentclass[sort&compress]{elsarticle}
\usepackage{blindtext, rotating}
\usepackage{color}
\usepackage[figure,vlined]{algorithm2e}
\SetCommentSty{textsf}
\SetKwRepeat{DoWhile}{do}{while}
\SetAlFnt{\small} 
\SetKw{Fn}{Procedure}
\newcommand{\Function}[2]{\Fn{\ProcNameSty{#1}}(#2)\par}
\usepackage{hyperref}
\usepackage{graphicx}
\usepackage{paralist}
\usepackage[english]{babel}

\newcommand{\commento}[1]{}
\newcommand{\algo}{\textsf{MagiCoder}}
\newcommand{\med}{\texttt{MedDRA}}
\newcommand{\vigi}{\texttt{VigiFarmaco}}
\newcommand{\adr}{ADR}
\newcommand{\llt}{\texttt{LLT}}
\newcommand{\pt}{\texttt{PT}}
\newcommand{\vllt}{\mathsf{Voted_{\llt}}}
\newcommand{\sllt}{\mathsf{Selected_{\llt}}}
\newcommand{\cone}[1]{\mathsf{C}_{1}(#1)}
\newcommand{\ctwo}[1]{\mathsf{C}_{2}(#1)}
\newcommand{\cthree}[1]{\mathsf{C}_{3}(#1)}
\newcommand{\cfour}[1]{\mathsf{C}_{4}(#1)}
\newcommand{\cfive}[1]{\mathsf{C}_{5}(#1)}

\newcommand{\svllt}{\mathsf{SortedVoted_{LLT}}}
\newcommand{\selvllt}{\mathsf{SelVoted_{LLT}}}
\newcommand{\finalvllt}{\mathsf{FinalVoted_{LLT}}}
\newcommand{\vigis}{\textsf{VigiSegn}}
\newcommand{\voters}[1]{\mathsf{voters}_{#1}}
\newcommand{\voted}[1]{\mathsf{voted}_{#1}}
\newcommand{\weights}[1]{\mathsf{weights}_{#1}}
\newcommand{\rels}{\mathsf{RelS}}
\newcommand{\rets}{\mathsf{RetS}}

\begin{document}
\title{From narrative descriptions to MedDRA:\\automagically encoding adverse drug reactions}

\author[cs]{Carlo Combi}
\ead{carlo.combi@univr.it}

\author[cs]{Margherita Zorzi}
\ead{margherita.zorzi@univr.it}

\author[dph]{Gabriele Pozzani}
\ead{gabriele.pozzani@univr.it}

\author[dph]{Ugo Moretti}
\ead{ugo.moretti@univr.it}

\address[cs]{Department of Computer Science, University of Verona, Italy}
\address[dph]{Department of Diagnostics And Public Health, University of Verona, Italy}

\begin{abstract}
\noindent\textit{Context}\hspace{0.5ex}
The collection of narrative spontaneous reports is an irreplaceable source for the prompt detection of suspected adverse drug reactions (ADRs): qualified domain experts manually revise a huge amount of narrative descriptions and then encode texts according to \med\ standard terminology.  The manual annotation  of narrative documents with medical terminology is a subtle and expensive task, since the number of reports is growing up day-by-day.

\noindent\textit{Objectives}\hspace{0.5ex} Natural Language applications can support the work of people responsible for pharmacovigilance. Our objective is to develop Natural Language Processing (NLP) algorithms and tools oriented to the healthcare domain,  in particular to the detection of ADR clinical terminology. Efficient applications can concretely improve the quality of the experts' revisions: NLP software can quickly analyze narrative texts  and offer a (as much as possible) correct solution (a list of \med\ terms) that the expert has to revise and validate.

\noindent\textit{Methods}\hspace{0.5ex} \algo, a  Natural Language Processing algorithm, is proposed for the automatic encoding of free-text descriptions into \med\ terms.  \algo\ procedure is  efficient in terms of computational complexity (in particular, it is linear in the size of the narrative input and the terminology). We tested it on a large dataset of about 4500 manually revised reports, by performing an automated comparison between human and \algo\ revisions. 

\noindent\textit{Results}\hspace{0.5ex} For the current base version of \algo, we measured: on short descriptions, an average recall of $86\%$ and an average precision of $88\%$; on medium-long descriptions (up to 255 characters), an average recall of $64\%$ and an average precision of $63\%$.

\noindent\textit{Conclusions}\hspace{0.5ex} 
From a practical point of view, \algo{} reduces the time required for encoding ADR reports. Pharmacologists have simply to review and validate the \med{} terms proposed by the application, instead of choosing the right terms among the 70K low level terms of \med{}. Such improvement in the efficiency of pharmacologists' work has a relevant impact also on the quality of the subsequent data analysis.
We developed \algo{} for the Italian pharmacovigilance language. However, our proposal is based on a general approach, not depending on the considered language nor the term dictionary.

\end{abstract}


\maketitle

\section{Introduction}
\label{section-introduction}

Pharmacovigilance includes all activities aimed to systematically study risks and benefits related to the correct use of marketed drugs.
The development of a new drug, which begins with the production and ends with the commercialization of a pharmaceutical product, considers both pre-clinical studies (usually  tests on  animals) and clinical studies (tests on patients).
After these phases, a pharmaceutical company can require the authorization for the commercialization of the new drug. Notwithstanding, whereas at this stage drug benefits are well-know, results about drug safety are not conclusive~\cite{SafetyDrug}.
The pre-marketing tests cited above have some limitations: they involve a small number of patients; they exclude relevant subgroups of population such as children and elders; the experimentation period is relatively short, less than two years; the experimentation does not deal with possibly concomitant pathologies, or with the concurrent use of other drugs.
For all these reasons, non-common Adverse Drug Reactions (ADRs), such as slowly-developing pathologies (e.g., carcinogenesis) or pathologies related to specific groups of patients, are hardly discovered before the commercialization. It may happen that drugs are withdrawn from the market after the detection of unexpected collateral effects.
Thus, it stands to reason that the post-marketing control of ADRs is a necessity, considering the mass production of drugs. As a consequence, pharmacovigilance plays a crucial role in human healthcare improvement~\cite{SafetyDrug}. 

Spontaneous reporting is the main method pharmacovigilance adopts in order to identify adverse drug reactions.
Through spontaneous reporting, health care professionals, patients, and pharmaceutical companies can voluntarily send information about suspected ADRs to the national regulatory authority\footnote{in Italy, the Drug Italian Agency AIFA -- Agenzia Italiana del FArmaco, http://www.agenziafarmaco.gov.it/}.
The spontaneous reporting is an important activity. It provides pharmacologists and regulatory authorities with early alerts, by considering every drug on the market and every patient category.

\begin{figure} 
\centering
\includegraphics[scale=.4]{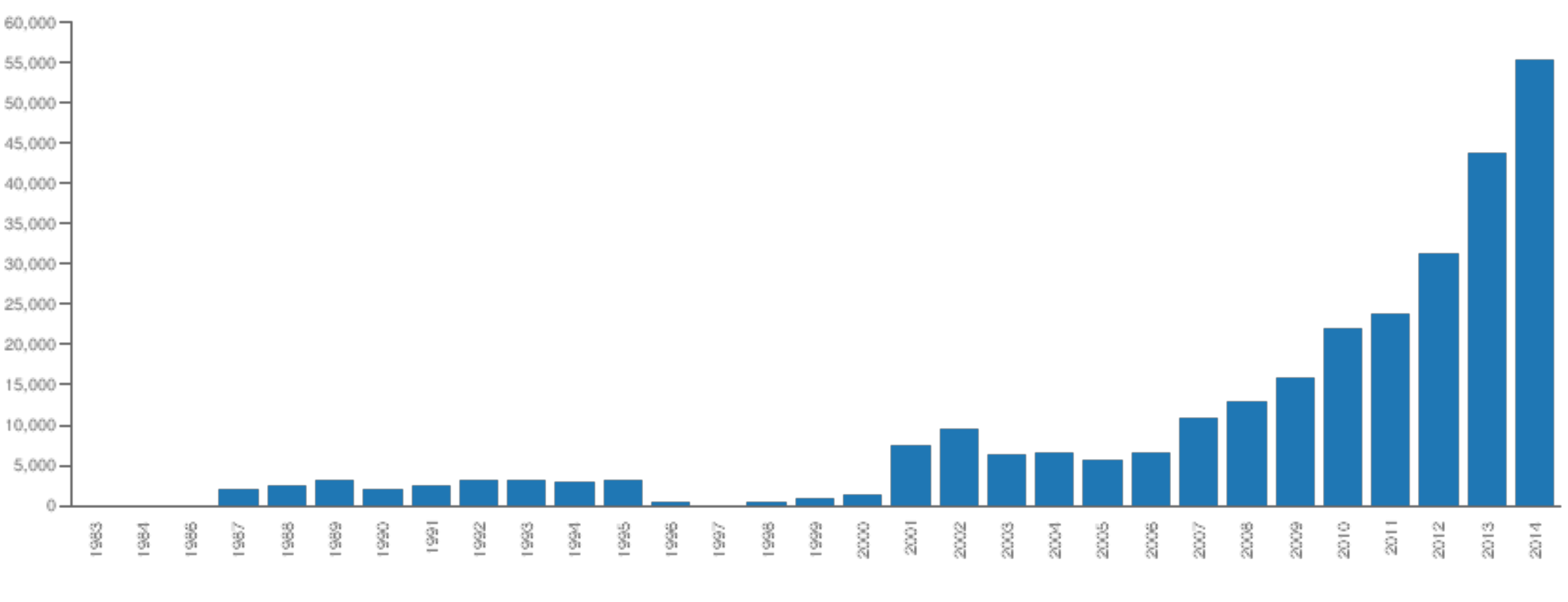}
\caption{The yearly increasing number of reports about suspected adverse reactions induced by drugs in Italy.}
\label{fig:Increasing-of-reports}
\end{figure}

The Italian system of pharmacovigilance requires that in each local healthcare structure (about 320 in Italy) there is a qualified person responsible for pharmacovigilance. Her/his assignment is to collect reports of suspected ADRs and to send them to the National Network of Pharmacovigilance (RNF, in Italian) within seven days since they have been received\footnote{According to the Italian Law, Art. 132 of Legislative Decree Number 219 of 04/24/2006.}.
Once reports have been notified and sent to RNF they are analysed by both local pharmacovigilance centres and by the Drug Italian Agency (AIFA). Subsequently, they are sent to Eudravigilance~\cite{Borg11} and to VigiBase~\cite{Aagard12} (the European and the worldwide pharmacovigilance network RNF is part of, respectively).
In general, spontaneous ADR reports are filled out by health care professionals (e.g., medical specialists, general practitioners, nurses), but also by citizens.
In last years, the number of ADR reports in Italy has grown rapidly, going from approximately ten thousand in 2006 to around sixty thousand in 2014~\cite{Vigisegn}, as shown in Figure~\ref{fig:Increasing-of-reports}.

Since the post-marketing surveillance of drugs is of paramount importance, such an increase is certainly positive. At the same time, the manual review of the reports became difficult and often unbearable both by people responsible for pharmacovigilance and by regional centres.
Indeed, each report must be checked, in order to control its quality; it is consequently encoded and transferred to RNF via ``copy by hand'' (actually, a printed copy).

Recently, to  increase the efficiency in collecting and managing ADR reports, a web application, called \vigi{}\footnote{Available at https://www.vigifarmaco.it}, has been designed and implemented for the Italian pharmacovigilance. 
Through \vigi, a spontaneous report can be filled out online by both healthcare professionals and citizens (through different user-friendly forms), as anonymous or registered users. The user is guided in compiling the report, since it has to be filled step-by-step (each phase corresponds to a different report section, i.e., ``Patient'', ``Adverse Drug Reaction'', ``Drug Treatments'', and ``Reporter'', respectively). At each step, data are validated and only when all of them have been correctly inserted the report can be successfully submitted.

Once ADR reports are submitted, they need to be validated by a pharmacovigilance supervisor. \vigi\ provides support also in this phase and is useful also for pharmacovigilance supervisors. Indeed, \vigi\ reports are high-quality documents, since they are automatically validated (the presence, the format, and the consistency of data are validated at the filling time). As a consequence, they are easier to review (especially with respect to printed reports). Moreover, thanks to \vigi, pharmacologists can send reports (actually, XML files~\cite{dexa15}) to RNF by simply clicking a button, after reviewing it.

Online reports have grown up to become the 30\% of the total number of Italian reports. As expected, it has been possible to observe that the average time between the dispatch of online reports and the insertion into RNF is sensibly shorter with respect to the insertion from printed reports.
Notwithstanding, there is an operation which still requires the manual intervention of responsibles for pharmacovigilance also for online report revisions: the encoding in \med\ terminology of the free text, through which the reporter describes one or more  adverse drug reactions. \med\ (Medical Dictionary for Regulatory Activities) is a medical terminology introduced with the purpose to standardize and facilitate the sharing of information about medicinal products in particular with respect to regulatory activities \cite{MeddraGuide}.
The description of a suspected ADR through narrative text could seem redundant/useless. Indeed, one could reasonably imagine sound solutions based either on an autocompletion form or on a menu with \med\ terms. In these solutions, the description of ADRs would be directly encoded by the reporter and no expert work for \med\ terminology extraction would be required. 
However, such solutions are not completely suited for the pharmacovigilance domain and the narrative description of ADRs remains a desirable feature, for at least two reasons. First, the description of an ADR by means of one of the seventy thousand \med\ terms is  a complex task. In most cases, the reporter who points out the adverse reaction is not an expert in \med\ terminology. This holds in particular for citizens, but it is still valid for several professionals. Thus, describing ADRs  by means of natural language sentences is simpler.
Second, the choice of the suitable term(s) from a given list or from an autocompletion field can influence the reporter and limit her/his expressiveness. As a consequence, the quality of the description would be also in this case undermined.
Therefore, \vigi\ offers a free-text field for specifying the ADR with all the possible details, without any restriction about the content or strict limits to the length of the written text. Consequently, \med\ encoding has then to be manually implemented by qualified people responsible for pharmacovigilance, before the transmission to RNF. As this work is expensive in terms of time and attention required, a problem about the accuracy of the encoding may occur given the continuous growing of the number of reports. 

According to the described scenario, in this paper we propose $\algo$, an \emph{original} Natural Language Processing (NLP)~\cite{Jurafsky2000} algorithm and related software tool, which automatically assigns one or more terms from a dictionary to a narrative text.
A preliminary version of $\algo$ has been proposed in~\cite{ZorziCLPM15}. \algo{} has been first developed for supporting pharmacovigilance supervisors in using \vigi{}, providing them with an initial automatic \med{} encoding of the ADR descriptions in the online reports collected by \vigi, that the supervisors check and may correct or accept as it is. In this way, the encoding task, previously completely manual, becomes semi-automatic, reducing errors and the required time for accomplishing it. In spite of its first goal, \algo{} has now evolved in an autonomous algorithm and software usable in all contexts where terms from a dictionary have to be recognized in a free narrative text. With respect to other solutions already available in literature and market, \algo{} has been designed to be efficient and less computationally expensive, unsupervised, and with no need of training. \algo{} uses stemming to be independent from singular/plural and masculine/feminine forms. Moreover, it uses string distance and other techniques to find best matching terms, discarding similar and non optimal terms.

With respect to the first version~\cite{ZorziCLPM15}, we extended our proposal following several directions. First of all, we refined the procedure: \algo\ has been equipped with some heuristic criteria and we started to address the problem of including auxiliary dictionaries (e.g., in order to deal with synonyms). \algo\ computational complexity has been carefully studied and we will show that it is linear in the size of the dictionary (in this case, the number of LLTs in \med) and the text description. We performed an accurate test of \algo\ performances: by means of well-known statistical measures, we collected a significant set of quantitative information about the effective behavior of the procedure. We largely discuss some crucial key-points we met in the development of this version of \algo, proposing short-time solutions we are addressing as work in progress, such as changes in stemming algorithm, considering synonyms, term filtering heuristics.

The paper is organized as follows. In Section~\ref{sec:related} we provide some background notions and we discuss related work. 
In Section~\ref{sec:magicoder} we present the algorithm \algo, by providing both a qualitative description and the pseudocode. In Section~\ref{sec:uiandbenchmark} we spend some words about the user interface of the related software tool. In Section~\ref{sec:testing} we explain the benchmark we developed to test $\algo$ performances and its results. Section~\ref{sec:cons} is devoted to some discussions. Finally, in Section~\ref{sec:future} we summarize the main features of our work and sketch some future research lines.

\section{Background and related work}\label{sec:related}

\subsection{Natural language processing and text mining in medicine}


Automatic detection of adverse drug reactions from text has recently received an increasing interest in pharmacovigilance research. 
Narrative descriptions of ADRs come from heterogeneous sources: spontaneous reporting, Electronic Health Records, Clinical Reports, and social media.  
In~\cite{Bate,Wang09,Friedman,Aramaki10,Dunagan08} some NLP approaches have been proposed for the extraction of ADRs from text. 
In~\cite{Bailey09}, the authors collect narrative discharge summaries from the Clinical Information System at New York Presbyterian Hospital. MedLEE, an NLP system, is applied to this collection, for identifing medication events and entities, which could be potential adverse drug events. Co-occurrence statistics with adjusted volume tests were used to detect associations between the two types of entities, to calculate the strengths of the associations, and to determine their cutoff thresholds. 
In~\cite{Toldo12},  the authors report on the adaptation of a machine learning-based system for the identification and extraction of ADRs in case reports.
The role of NLP approaches in optimised machine learning algorithms is also explored in~\cite{Gonzalez14}, where the authors address the problem of  automatic detection of ADR assertive text segments from several sources, focusing on data posted by users on social media (Twitter and DailyStrenght, a health care oriented social media).  Existing methodologies for NLP are discussed and an experimental comparison between NLP-based machine learning algorithms over data sets from different sources is proposed. Moreover, the authors address the issue of data imbalance for ADR description task. In~\cite{Yang12} the authors propose to use association mining and Proportional Reporting Ratio (PRR, a well-know pharmacovigilance statistical index) to mine the associations between drugs and adverse reactions from the user contributed content in social media. In order to extract adverse reactions from on-line text (from health care communities), the authors apply the Consumer Health Vocabulary\footnote{Available at http://www.consumerhealthvocab.org} to generate ADR lexicon. ADR lexicon is a computerized collection of health expressions derived from actual consumer utterances, linked to professional concepts and reviewed and validated by professionals and consumers. Narrative text is preprocessed following standard NLP techniques (such as stop word removal, see Section~\ref{sec:description}). An experiment using ten drugs and five adverse drug reactions is proposed. The Food and Drug Administration alerts are used as the gold standard, to test the performance of the proposed techniques. 
The authors developed algorithms to identify ADRs from threads of drugs, and implemented association mining to calculate leverage and lift for each possible pair of drugs and adverse reactions in the dataset. At the same time, PRR is also calculated.

Other related papers about pharmacovigilance and machine learning or data mining are~\cite{Harpaz10,Bellazzi15}. 
In~\cite{Uppsala}, a text extraction tool is implemented on the .NET platform for preprocessing text (removal of stop words, Porter stemming~\cite{Porter80} and use of synonyms) and matching medical terms using permutations of words and spelling variations (Soundex, Levenshtein distance and Longest common subsequence distance~\cite{Collins03}). Its performance has been evaluated on both manually extracted medical terms from summaries of product characteristics and unstructured adverse effect texts from Martindale (a medical reference for information about drugs and medicines) using the WHO-ART and \med\ medical terminologies. A lot of linguistic features have been considered and a careful analysis of performances has been provided.
In~\cite{Coff07} the authors develop an algorithm in order to help coders in the subtle task of auto-assigning ICD-9 codes to clinical narrative descriptions. Similarly to \algo, input descriptions are proposed as free text. The test experiment takes into account a reasoned data set of manually annotated radiology reports, chosen to cover all coding classes according to ICD-9 hierarchy and classification: the test obtains an accuracy of $77\%$.

\subsection{\med\ Dictionary}

The Medical Dictionary for Regulatory Activities (\med) \cite{MeddraGuide} is a medical terminology used to classify adverse event information associated with the use of biopharmaceuticals and other medical products (e.g., medical devices and vaccines). Coding these data to a standard set of MedDRA terms allows health authorities and the biopharmaceutical industry to exchange and analyze data related to the safe use of medical products~\cite{Radhakrishna14}. It has been developed by the International Conference on Harmonization (ICH); it belongs to the International Federation of Pharmaceutical Manufacturers and Associations (IFPMA); it is controlled and periodically revised by the \med\ Mainteinance And Service Organization (MSSO).
\med\ is available in eleven European languages and in Chinese and Japanese too. It is updated twice a year (in March and in September), following a collaboration-based approach: everyone can propose new reasonable updates or changes (due to effects of events as the onset of new pathologies) and a team of experts eventually decides about the publication of updates.
\med{} terms are organised into a hierarchy: 
the SOC (System Organ Class) level includes the most general terms; 
the LLT (Low Level Terms) level includes more specific terminologies. Between SOC and LLT there are three intermediate levels: HLGT (High Level Group Terms), HLT (High Level Terms), and PT (Preferred Terms).

The encoding of ADRs through \med\ is extremely important for report analysis as for a prompt detection of problems related to drug-based treatments.
Thanks to \med\, it is possible to group similar/analogous cases described in different ways (e.g., by synonyms) or with different details/levels of abstraction.

Table~\ref{tab:exmeddra} shows an example of the hierarchy: reaction \emph{Itch} is described starting from \emph{Skin disorders} (SOC), \emph{Epidermal conditions} (HLGT), \emph{Dermatitis and Eczema} (HLT), and \emph{Asteatotic Eczema} (PT). Preferred Terms are Low Level Terms chosen to be representative of a group of terms. It should be stressed that the hierarchy is multiaxial: for example, a PT can be grouped into one or more HLT, but it belongs to only one primary SOC term.
 
\begin{table}
\centering
\begin{tabular}{|c|c|}
  \hline
  \rule{0em}{1.1em}\med\ Level & \med\ Term     \\
  \hline
  \rule{0em}{1.1em}SOC & Skin disorders           \\
	\hline
  \rule{0em}{1.1em}HLGT & Epidermal conditions\\
	\hline
  \rule{0em}{1.1em}HLT & Dermatitis and Eczema \\
      \hline
  \rule{0em}{1.1em}PT & Asteatotic Eczema\\
	\hline
  \rule{0em}{1.1em}LLT & Itch\\
	\hline
\end{tabular}
\caption{MedDRA Hierarchy - an Example}\label{tab:exmeddra}
\end{table}

\section{\algo: an NLP software for ADR automatic encoding}\label{sec:magicoder}

A natural language ADR description is a completely free text. The user has no limitations, she/he can potentially write everything: a number of online ADR descriptions actually contain information not directly related to drug effects. 
Thus, an NLP software has to face and solve many issues: Trivial orthographical errors; Use of singular versus plural nouns; The so called ``false positives'',  i.e., syntactically retrieved inappropriate results, which are closely resembling to correct solutions; The structure of the sentence, i.e., the way an assertion is built up in a given language.
Also the ``intelligent'' detection of linguistic connectives is a crucial issue. For example, the presence of a negation can potentially change the overall meaning of a description.

In general, a satisfactory automatic support of human reasoning and work is a subtle task: for example,  the uncontrolled extension of the dictionary with auxiliary synonymous (see Section~\ref{sec:syn}) or the naive ad hoc management of particular cases, can limit the efficiency and the desired of the algorithm. 
For these reasons, we carefully designed $\algo$, even through a side-by-side collaboration between pharmacologists and computer scientists, in order to yield an efficient tool, capable to really support pharmacovigilance activities.

In literature, several NLP algorithms already exist, and several interesting approaches (such as the so called morpho-analysis of natural language) have been studied and proposed~\cite{BauerLaurie2003,Jurafsky2000,Kishida2005}. According to the described pharmacovigilance domain, we considered algorithms for the morpho-analysis and the part-of-speech (PoS) extraction techniques~\cite{BauerLaurie2003,Jurafsky2000} too powerful and general purpose for the solution of our problem.  Indeed, in most cases ADR descriptions are written in a very succinct way, without using verbs, punctuation, or other lexical items, and introducing acronyms. Moreover, clinical and technical words are often not recognized correctly because not included in usual dictionaries. All these considerations limit the benefits of using morpho-analysis and PoS for our purposes.

Thus, we decided to design and develop an ad hoc algorithm for the problem we are facing, namely that of deriving \med\ terms from narrative text and mapping segments of text in effective \llt{}s. This task has to be done in a very feasible time (we want that each interaction user/\algo\ requires less than a second) and the solution offered to the expert has to be readable and useful. Therefore, we decided to ignore the \emph{structure} of the narrative description and address the issue in a simpler way.
Main features of \algo\ can be summarized as follows:
\begin{itemize}
\item it requires \emph{a single linear scan of the narrative description}: as a consequence, our solution is particularly efficient in terms of computational complexity;
\item it has been designed and developed for the specific problem of mapping Italian text to \med\ dictionary, but we claim the way \algo\ has been developed is sound with respect to language and dictionary changes (see Section~\ref{sec:future});
\item the current version of \algo\ is only based on the pure syntactical recognition of the text and it does not exploit any external synonym dictionary; in Section~\ref{sec:uiandbenchmark} we will discuss how synonyms may be used to increase \algo{} performances. In particular, we will discuss how a na\"ive approach to synonyms worsen computational and retrieval performances, while we will show through experimental results and empirical observations that a prudent and suitable use of an external dictionary produces an improvement of performances.
\end{itemize}

In this paper we consider the Italian context of Pharmacovigilance and, as a consequence, we will consider and process by \algo\ textual descriptions written in Italian language. We will discuss the potentiality of \algo\ on other languages and some preliminary results in Section~\ref{sec:future}. 

\subsection{\algo: overview}\label{sec:description}

The  main idea of $\algo$ is that a single linear scan of the free-text is sufficient, in order  to recognize $\med$ terms. 

From an abstract point of view, we try to recognize, in the narrative description, \emph{single words} belonging to \llt{}s, which do not necessarily occupy consecutive positions in the text. This way, we try to ``{reconstruct}'' \med\ terms, taking into account the fact that in a description the reporter can permute or omit words.
As we will show, \algo\ has not to deal with computationally expensive tasks, such as taking into account subroutines for permutations and combinations of words (as, for example, in~\cite{Uppsala}). 

We can distinguish  five phases in the procedure that will be discussed in detail in Sections~\ref{sec:dsd}, \ref{sec:preproc}, \ref{voting}, \ref{sec:criteriasorting}, \ref{sec:winning}, respectively.
\begin{enumerate}
\item{Definition of ad hoc data structures}: the design of data structures is central to perform an efficient computation; our main data structures are hash tables, in order to guarantee an efficient access both to \med\ terms and to words {belonging to} \med\ terms.
\item Preprocessing of the original text: tokenization (i.e., segmentation of the text into syntactical units), stemming (i.e., reduction of words to a particular root form), elimination of computationally irrelevant words.
\item Word-by-word linear scan of the description and ``voting task'':  a word ``votes'' \llt{}s it belongs to. For each term voted by one or more words, we store some information about the retrieved syntactical matching.
\item Weights calculation: recognized terms are weighted depending on information about syntactical matching.
\item  Sorting of voted terms and winning terms release: the set of voted term is pruned, terms are sorted and finally a solution (a set of winning terms) is released.
\end{enumerate}

\subsubsection{Definition of ad hoc data structures}\label{sec:dsd}

The algorithm proceeds with a word-by-word comparison. We iterate on the preprocessed text and we test if a single word $w$, a token, occurs into one or many \llt{}s.

In order to efficiently test if a token belongs to one or more \llt{}s, we need to know which words belong to each term.
{The \llt\ level of \med\ is actually a set of \emph{phrases}, i.e., \emph{sequences of words}. By scanning these sequences, we build a \emph{meta-dictionary} of all the words which compose  \llt{}s.}  
As we will describe in Section~\ref{sec:sketchcompl}, in $O(mk)$ time units (where $m$ and $k$ are the cardinality of the set of \llt{}s and the length of the longest \llt\ in \med, respectively) we build a hash table having all the words occurring in \med\ as keys, where the value associated to key $w_i$ contains information about the set of  \llt s containing $w_i$.
This way, we can verify the presence in \med\ of a word $w$ encountered in the \adr\ description in constant time.  We call this meta-dictionary $\mathsf{DictByWord}$.
We build a meta dictionary also from a stemmed version of \med, to verify the presence of stemmed descriptions. We call it $\mathsf{DictByWordStem}$. Finally, also the \med\ dictionary is loaded into a hash table according to LLT identifiers and, in general, all our main data structures are hash tables.

We aim to stress that, to retain efficiency, we preferred exact string matching with respect to approximate string matching, when looking for a word into the meta dictionary.  Approximate string matching would allow us to retrieve terms that would be lost in exact string matching (e.g., we could recognize misspelled words in the ADR description), but it would worsen the performances of the text recognition tool, since direct access to the dictionary would not be possible. We discuss the problem of retrieving syntactical variations of the same words and the problem of addressing orthographical errors in Section~\ref{sec:future}. 

\subsubsection{Preprocessing of the original \adr\ description}\label{sec:preproc}

Given a natural language \adr\ description, the text has to be preprocessed in order to perform an efficient computation. We adopt a well-know technique such as tokenization~\cite{Clark2010}: a phrase is reduced to \emph{tokens}, i.e., syntactical units which often, as in our case, correspond to words. A tokenized text can be easily manipulated as an enumerable object, e.g., an array.
A \emph{stop word} is a word that can be considered irrelevant for the text analysis (e.g., an article or an interjection). Words classified as stop-words are removed from the tokenized text.
In particular, in this release of our software we decided to not take into account \emph{connectives}, e.g., conjunctions, disjunctions, negations. The role of connectives, in particular of negation, is discussed in Section~\ref{sec:cons}.

A fruitful preliminary work is the extraction of the corresponding \emph{stemmed} version from the original tokenized and stop-word free text. 
Stemming is a linguistic technique that, given a word, reduces it to a particular kind of root form~\cite{Porter80,Clark2010}. It is useful in text analysis, in order to avoid problems such as missing word recognition due to singular/plural forms (e.g., hand/hands). In some cases, stemming procedures are able to recognize the same root both for the adjectival and the noun form of a word. Stemming is also potentially harmful, since it can generate so called ``false positives'' terms. A meaningful example can be found in Italian language. The  plural of the word \emph{mano} (in English, \emph{hand}) is \emph{mani} (in English, \emph{hands}), and their stemmed root is \emph{man}, which is also the stemmed version of \emph{mania} (in English, \emph{mania}).
Several stemming algorithms exist, and their impact on the performances of \algo\ is discussed in Section~\ref{sec:cons}.

\subsubsection{Word-by-word linear scan of the description and voting task}\label{voting}
 
$\algo$ scans the text word-by-word (remember that each word corresponds to a token) once and performs a ``voting task'': at the $i$-th step,  it marks (i.e., ``votes'')  with index $i$ each \llt\ $t$ containing the current ($i$-th) word of the \adr\ description.  Moreover, it keeps track of the position where the $i$-th word occurs in $t$.

$\algo$ tries to find a word match both for the exact and the stemmed version of the meta dictionary and keeps track of the kind of match it has eventually found. It updates a flag, initially set to 0, if at least a stemmed matching is found in an \llt. If a word $w$ has been exactly recognized in a term $t$, the  match between the stemmed versions of $w$ and $t$ is not considered.
At the end of the scan, the procedure has built a sub-dictionary containing only terms ``voted'' at least by one word. We call $\vllt$ the sub-dictionary of voted terms. 

Each voted term $t$ is equipped with two auxiliary data structures, containing, respectively:

\begin{enumerate}
\item the positions of the \emph{voting words} in the \adr\ description; we call $\voters{t}$ this sequence of indexes;
\item the positions of the \emph{voted words} in the \med\ term $t$; we call $\voted{t}$ this sequence of indexes.
\end{enumerate}

Moreover, we endow each voted term $t$ with a third structure that will contain the \emph{sorting criteria} we define below; we will call it $\weights{t}$. 

Let us now introduce some notations we will use in the following. We denote as $t.size$ the function that, given an \llt\ $t$, returns the number of words contained in $t$ (excluding the stop words). We denote as $\voters{t}.length$ (resp. $\voted{t}.length$) the function that returns the number of indexes belonging to $\voters{t}$ (resp. $\voted{t}$). 
We denote as $\voters{t}.min$  and $\voters{t}.max$ the functions that return the maximum and the minimum indexes in  $\voters{t}$, respectively.

From now on,  sometimes we explicitly list the complete  denomination of a terms: we will use the notation \emph{``name''(id)}, where \emph{``name''} is the  \med\ description and \emph{id} is its identifier, that is possibly used to refer to the term.
Let us exemplify these notions by introducing an example. Consider the following ADR description: ``anaphylactic shock (hypotension + cutaneous rash) 1 hour after
taking the drug''. Words in it are numbered from 0 (anaphylactic) to 9 (drug). The complete set of data structures coming from the task is too big to be reported here, thus we focus only on two \llt{}s. At the end of the voting task, $\vllt$ will include, among others, ``Anaphylactic shock'' (10002199) and ``Anaphylactic reaction to drug'' (10054844). We will have that $\voters{10002199}=[0, 1]$ (i.e., ``anaphylactic'' and ``shock'') while $\voters{10054844}=[0, 9]$ (i.e., ``anaphylactic'' and ``drug''). On the other hand, $\voted{10002199}=[0,1]$, revealing that both words in the term have been voted, while $\voted{10054844}=[0,2]$, suggesting that only two out of three words in the term have been voted (in particular, ``reaction'' has not been voted). In this example all words in the description have been voted without using the stemming.

\subsubsection{Weight calculation}\label{sec:criteriasorting}

After the voting task, selected terms have to be ordered. Notice that a purely syntactical recognition of words in \llt{}s potentially generates a large number of voted terms. For example, in the Italian version of \med, the word ``male'' (in English, ``pain'') occurs 3385 times. 

So we have to: i) filter a subset of highly feasible solutions, by means of quantitative weights we assigns to candidate solutions; ii) choose a good final selection strategy in order to release a small set of final ``winning'' \med\ terms (this latter point will be discussed in Section~\ref{sec:winning}).

For this purpose, we define four criteria to assign ``weights'' to voted terms accordingly.

In the following, $\frac{1}{t.size}$ is a normalization factor (w.r.t. the length, in terms of words, of the \llt\ $t$). First three criteria have 0 as optimum value and 1 as worst value, while the fourth criterion has optimum value to 1 and it grows in worst cases.

\begin{description}
\item [\textbf{Criterion one: Coverage}]\mbox{}

First, we consider how much part of the words of each voted \llt\ have not been recognized.

\begin{displaymath}
\cone{t}=\frac{t.size-\voted{t}.length}{t.size}
\end{displaymath}

In the example we introduced before, we have that $\cone{10002199}=0$ (i.e., all words of the terms have been recognized in the description) while $\cone{10054844}=0.33$ (i.e., one word out of three has not been recognized in the description).

\item [\textbf{Criterion two: Type of Coverage}]\mbox{}

The algorithm considers whether a perfect matching has been performed using or not stemmed words. $\ctwo{\cdot}$ is simply a flag. $\ctwo{t}$ holds if stemming has been used at least once in the voting procedure of $t$, and it is valued 1, otherwise it is valued 0.

For example, $\ctwo{10002199}=0$ and $\ctwo{10054844}=0$.

\item [\textbf{Criterion three: Coverage Distance}]\mbox{}

The use of stemming allows one to find a number of (otherwise lost) matches. As side effect, we often obtain a quite large set of joint winner candidate terms. In this phase, we introduce a string distance comparison between recognized words in the original text and voted \llt{}s. Among the possible string metrics, we use the so called pair distance~\cite{string}, which is robust with respect to word permutation. Thus,

\begin{displaymath}
\cthree{t}=\mathit{pair}(t,\overline{t})
\end{displaymath}

where $\mathit{pair}(s,r)$ is the pair distance function (between strings $s$ and $r$) and $\overline{t}$ is the term ``rebuilt'' from the words in \adr\ description corresponding to indexes in $\voters{t}$.

For example, $\cthree{10002199}=0$ (i.e., the concatenation of the voters and the term are equal) and $\cthree{10054844}=12$.

\item [\textbf{Criterion four: Coverage Density}]\mbox{}

We want to estimate how an \llt\ has been covered. 

\begin{displaymath}
\cfour{t}=\frac{(\voters{t}.max-\voters{t}.min)+1}{\voted{t}.length}
\end{displaymath}

The intuitive meaning of the criterion is to quantify the ``quality'' of the coverage. If an \llt\ has been covered by nearby words, it will be considered a good candidate for the solution. This criterion has to be carefully implemented, taking into account possible duplicated voted words.
\end{description}

After computing (and storing) the weights related to the above criteria, for each voted term $t$ we have the data structure $\weights{t}=[\cone{t}, \ctwo{t}, \cthree{t}, \cfour{t}]$, containing the weights corresponding to the four criteria.
These weights will be used, after a first heuristic selection, to sort a subset of the syntactically retrieved terms. 

Continuing the example introduced before, we have that $\cfour{10002199}=1$ while $\cfour{10054844}=5$. Thus, concluding, we obtain that $\weights{10002199} = [0, 0, 0, 1]$ while $\weights{10054844} = [0.33, 0, 12, 5]$.

\subsubsection{Selection, ordering and release of winning terms}\label{sec:winning}

In order to provide an effective support to pharmacovigilance experts' work, it is important to offer only 
a small set of good candidate solutions. 

As previously said, the pure syntactical recognition of \med\ terms into a free-text generates a possibly large set of results. 
Therefore, the releasing strategy has to be carefully designed in order to select onlt best suitable solutions. We will provide an heuristic selection, followed by a sorting of the survived voted terms;  then we propose a release phase of solutions, further refined by a final heuristic criterium. 

\medskip

As a first step, we provide an initial pruning of the syntactically retrieved terms guided by the \emph{ordered-phrases}  heuristic criterium.
In the ordered-phrases criterium we reintroduce the order of words in the narrative description as a selection discriminating factor. From the set of selected \llt{}s, we remove those terms where voters (i.e., tokens in the original free text) appear in the ADR description in a relative order different from that of the corresponing voted tokens in the LLT. We do that \emph{only} for those \llt{}s having voters that voted for more than one term.

Let us consider the following example. On the (Italian) narrative description ``edema della glottide-lingua, parestesia al volto, dispnea'' (in English, ``edema glottis-tongue, facial paresthesia, dyspnoea''), the voting procedure of \algo\ finds, among the solutions, the \med\ terms ``Edema della glottide'' (``Edema glottis''), ``Edema della lingua'' (``Edema tongue''), ``Edema del volto'' (``Edema face''), ``Parestesia della lingua'' (``Paresthesia tongue''), and ``Dispnea'' (``Dyspnoea''). The ordererd-phrase criterium removes \llt{} ``Parestesia della lingua'' from the set of candidate solutions because ``lingua'' votes for two terms but in the narrative text it appears before than ``parestesia'' while in the \llt{} it appears after.
\medskip

We call $\selvllt$ the set of voted terms after the selection by the ordered-phrases criterium.
We proceed then by ordering $\selvllt$: we use a multiple-value sorting on elements in $\weights{t}$, for each $t\in\selvllt$. The obtained subdictionary is dubbed as $\svllt$ and it has possibly most suitable solutions on top.
\medskip

After this phase, the selection of the ``winning terms'' takes place.
The main idea is to select and return a subset of voted terms which ``covers'' the \adr\ description.
We create the set $\sllt$ as follows.
We iterate on the ordered dictionary and for each $t \in \svllt$ we select $t$ if all the following conditions hold: 
\begin{enumerate}
\item $t$ is completely covered, i.e., $\cone{t}=0$;
\item $t$ does not already belong to $\sllt$; 
\item $t$ is not a prefix of another selected term $t' \in \svllt$;
\item $t$ has been voted without stemming (i.e., $\ctwo{t}=0$) or, for any $w_i \in \voters{t}$,  
$w_i$ has not been covered (i.e., none term voted by $w_i$ has been already selected) or $w_i$ has not been exactly covered (i.e., only its stem has been recognized in some term $t_1$)\footnote{In the implementation we add also the following thresholds: we choose only terms $t$ such that $\cthree{t}<0.5$ and $\cfour{t}<3$. We extracted these thresholds by means of some empirical tests. We plan to eventually adjust them after some further performance tests.}.
\end{enumerate}


At this stage, we have a set of \med\ terms which ``covers'' the narrative description. 
We further select a subset $\finalvllt$ of $\sllt$ with a second heuristic, the \emph{maximal-set-of-voters} criterium. 

The maximal-set-of-voters criterium deletes from the solution those terms which can be considered ``extensions'' of other ones. For each pair of terms $t_i$ and $t_j$, it checks if $\voters{t_i}$ is a subset of $\voters{t_j}$ (considered as sets of indexes). If it is the case, $t_i$ is removed from $\sllt$.

In $\algo$ we do not need to consider ad hoc subroutines to address permutations and combinations of words (as it is done, for example, in~\cite{Uppsala}). In Natural Language Processing, permutations and combinations of words are important, since in spoken language the order of words can change w.r.t. the formal structure of the sentences. Moreover, some words can be omitted, while the sentence still retains the same meaning.  These aspects come for free from our voting procedure: after the scan, we retrieve the information that \emph{a set of words covers a term} $t\in\vllt$, but {the order between words does not necessarily matter.}

\subsection{\algo: structure of the algorithm}\label{sec:pseudocode}

Figure~\ref{tab:pseudocode} depicts the pseudocode of \algo. We represent dictionaries  either as sets of words or as sets of functions. 
We describe the main procedures and functions used in the pseudocode.
\begin{itemize}
\item Procedure $Preprocessing$ takes the narrative description, performs tokenization and stop-word removal and puts it into an array of words.
\item Procedures $CreateMetaDict$ and $CreateMetaDictStem$ get \llt{}s and create a dictionary of \emph{words} and of their stemmed versions, respectively, which belong to \llt{}s, retaining the information about the set of terms containing each word.
\item By the functional notation $\mathsf{DictByWord}(w)$ (resp., $\mathsf{DictByWordStem}(w)$), we refer to the set of \llt{}s containing the word $w$ (resp., the stem of $w$).
\item Function $stem(w)$ returns the stemmed version of word $w$. 
\item Function $indx_{t}(w)$ returns the position of word $w$ in term $t$. 
\item $stem\_usage_{t}$ is a flag, initially set to 0, which holds 1 if at least a stemmed matching  with the \med\ term $t$ is found.
\item $\mathsf{adr\_clear}$, $\voters{t}$, $\voted{t}$ are arrays and  $\mathsf{add}[A,l]$ appends $l$ to array $A$, where $l$ may be an element or a sequence of elements. 

\item $\mathsf{C}_{i}$ ($i=1,2,3,4$) are the weights related to the criteria defined in Section~\ref{sec:criteriasorting}.
\item Procedure $sortby(A, \{v_1,\ldots,v_k\})$ performs the multi-value sorting of the array $A$ based on the values of the properties $v_1,\ldots,v_k$ of its elements.
\item Procedure $prefix(S,t)$, where $S$ is a set of terms and $t$ is a term, tests whether $t$ (considered as a string) is \emph{prefix} of a term in $S$. Dually,  procedure $remove\_prefix(S,t)$  tests if in $S$ there are one or more prefixes of $t$, and eventually remove them from $S$.
\item Function $mark(w)$ specifies whether a word $w$ has been already covered (i.e., a term voted by $w$ has been selected) in the (partial) solution during the term release:  $mark(w)$ holds 1 if $w$ has been covered (with or without stemming) and it holds 0 otherwise. We assume that before starting the final phase of building the solution (i.e., the returned set of \llt{}s), $mark(w)=0$ for any word $w$ belonging to the description.
\item Procedures $ordered\_phrases(S)$ and $maximal\_voters(S)$, where $S$ is a set of terms, implement  \emph{ordered-phrases} and \emph{maximal-set-of-voters} criteria (defined in Section~\ref{sec:winning}), respectively.
\item Function $win(S,n)$, returns the first $n$ elements of an ordered set $S$. If $|S|=m< n$, the function returns the complete list of ordered terms and $n-m$ nil values. 
\end{itemize}


\begin{algorithm*}[!t]
{\scriptsize
    \Function{\algo}{$D$ text, $\mathsf{LLTDict}$ dictionary, $n$ integer}
    \KwIn{$D$: the narrative description;\newline
      $\mathsf{LLTDict}$: a data structure containing the \med{} $\llt$s;\newline
      $n$: the maximum number of winning terms that have to be released by the procedure}
    \KwOut{an ordered set of \llt{}s}
     $\mathsf{DictByWord}$ = CreateMetaDict($\mathsf{LLTDict}$)\;
     $\mathsf{DictByWordStem}$ = CreateStemMetaDict($\mathsf{LLTDict}$)\;
     \emph{adr\_clear} = Preprocessing($D$)\;
     \emph{adr\_length} = \emph{adr\_clear}.length\;
    $\vllt$ = $\emptyset$\;
    \tcc{for each non-stop-word in the description}
    \ForEach{(i $\in [0, adr\_length-1]$ }{
    \tcc{test whether the current word belongs to  \med }
    		\If{adr\_clear[i] $\in \mathsf{DictByWord}$}{
        \tcc{for each term containing the word}
   			\ForEach{t $\in \mathsf{DictByWord}$(adr\_clear[i])}{
    			\tcc{keep track of the index of the voting word}
   			$\mathsf{add}$[$\voters{t}$,i]\;
           \tcc{keep track of the index of the recognized word in $t$ }
			$\mathsf{add}$[$\voted{t}$, $indx_{t}$(\textit{adr\_clear[i]})]\;

  			$\vllt$ = $\vllt \cup t$\;
   			}
   		}
      	\tcc{test if the current (stemmed) word belongs  the stemmed \med }
      	\If{stem({adr\_clear[i]}) $\in \mathsf{DictByWordStem}$}{
      		\ForEach{t $\in\mathsf{DictByWordStem}$(stem(adr\_clear[i]))}{
      			\tcc{ test if the current term has not been exactly voted by the same word }
 				\If{i $\notin \voters{t}$}{
 					$\mathsf{add}$[$\voters{t}$, i]\;
  					$\mathsf{add}$[$\voted{t}$, $indx_{t}$(\textit{adr\_clear[i]})]\;
     				\tcc{  keep track that $t$ has been covered by a stemmed word}
 					$stem\_usage_{t}$ = true\;
 				} 
  			$\vllt$ = $\vllt \cup t$
  			}
 		}
	}
	\tcc{ for each voted term, calculate the four weights of the corresponding criteria}
	\ForEach{t $\in\vllt$}{
		$\mathsf{add}$[$\weights{t}, \cone{t},\ctwo{t},\cthree{t},\cfour{t}$]
	}  
	\tcc{filtering of the voted terms by the first heuristic criterium}
	$\selvllt={orderd\_phrases}(\vllt)$\;
	\tcc{multiple value sorting of the voted terms}
	$\svllt$ = sortby($\selvllt, \{\mathsf{C}_{1},\mathsf{C}_{2},\mathsf{C}_{3},\mathsf{C}_{4}\}$)\;
	\ForEach{t $\in\svllt$}{
		\ForEach{index $\in\voters{t}$}{
			\tcc{select a term $t$ if it has been completely covered, its i-th voting word has not been covered or if its i-th voting word has been perfectly recognized in $t$ and if $t$ is not prefix of 
			another already selected terms}
			\If{$\mathsf{C}_{1}(t)=0$ AND (($stem\_usage_{t}$ = false OR (mark(adr\_clear(index))=0))
			AND t $\notin\sllt$ AND  prefix($\sllt$,t)=false)}{
				mark(adr\_clear(index))=1\;
				\tcc{remove from the selected term set all terms which are prefix of $t$}
				$\sllt$ = remove\_prefix($\sllt$,t)\;
				$\sllt$ = $\sllt \cup t$
			} 
		}
	}  
	\tcc{filtering of the finally selected terms by the second heuristic criterium}
	$\finalvllt ={maximal\_voters}(\sllt)$\;
	$\mathsf{winners}= win(\finalvllt,n)$\;
	\Return{$\mathsf{winners}$} 
}
	\caption{Pseudocode of \algo\ }\label{tab:pseudocode}
\end{algorithm*}



\subsection{\algo\ complexity analysis}\label{sec:sketchcompl}

Let us now conclude this section by sketching the analysis of the computational complexity of \algo.

Let $n$ be the input size (the length, in terms of words, of the narrative description). Let $m$ be the cardinality of the  dictionary (i.e., the number of terms). Moreover, let $m'$ be the number of distinct words occurring in the dictionary  and let $k$ be the length of the longest term in the dictionary.  For \med, we have about 75K terms ($m$) and 17K unique words ($m'$). Notice that, reasonably, $k$ is a small constant for any dictionary; in particular, for \med\ we have $k=22$. We assume that all update operations on auxiliary data structures require constant time $O(1)$.

Building  meta-dictionaries $\mathsf{DictByWord}$ and $\mathsf{DictByWordstems}$ requires $O(km)$ time units. In fact, the simplest procedure to build these hash tables is to scan the \llt\ dictionary and, for each term $t$, to verify for each word $w$ belonging to $t$ whether $w$ is a key in the hash table (this can be done in constant time). If $w$ is a key, then we have to update the values associated to $w$, i.e., we add $t$ to the set of terms containing $w$. Otherwise, we add the new  key $w$ and the associated term $t$ to the hash table. We note that these meta-dictionaries are computed only once when the \med{} dictionary changes (twice per year), then as many narrative texts as we want can be encoded without the need to rebuild them.

It can be easily verified that the voting procedure requires in the worst case $O(nm)$ steps: this is a totally conservative bound, since this worst case should imply that each word of the description appears in all the terms of the dictionary. A simple analysis of the occurrences of the words in \med{} shows that this worst case never occurs: in fact, the maximal absolute frequency of a \med\ word is  3937, and the average of the frequencies of the words is 19.1\footnote{These values have been calculated excluding the stop-words and taking into account the stems of the words appearing in \med{}.}. Thus, usually, real computational complexity is much less of this worst case.

The computation of criteria-related weights requires $O(nm)$ time units. In particular: both criterion one and criterion two require $O(m)$ time steps; criterion three require $O(nm)$ (we assume to absorb the complexity of the pair distance function); criterion four requires $O(nm)$ time units.

The subsequent multi-value sorting based on computed weights is a sorting algorithm which complexity can be approximated to $O(m \log{m})$, based on the comparison of objects of four elements (i.e., the weights of the four criteria). Since the number of the criteria-related weights involved in the multi-sorting is constant, it can be neglected. Thus, the complexity of multi-value sorting can be considered to be $O(m \log{m})$. 

Finally, to derive the best solutions actually requires $O(nm)$ steps. The ordered-phrases criterium requires $O(nm)$; the maximal set of voters criterium takes $O(mn)$ time units.

Thus, we conclude that \algo{} requires in the worst case $O(nm)$ computational steps. We again highlight that this is  a (very) worst case scenario, while in average it performs quite better. Moreover, we did not take into account that each phase works on a subset of terms of the previous phase, and the size of these subset rapidly decreases  in common application.

the selection phase works only on voted terms, thus, in common applications, on a subset of the original dictionary.

\commento{ 
\bigskip
\textcolor{blue}{Io non mi lancerei in questo confronto, a meno che gli autori da qualche parte non abbiano dato una stima formale della loro complessit\`a, devo verificare}
The computational complexity of \algo\ is likely to be lower than that of the tool proposed in~\cite{Uppsala}. Indeed, in \cite{Uppsala} the author describes a sophisticated procedure which considers also approximate string matching. This feature does not allow constant time search for text-dictionary  matches (i.e., it is not always possible to exploit direct data access through optimal data structures, such as hash tables). Moreover, explicitly considering word permutation and combination is a computationally  expensive task. We claim that the efficiency of \algo\ can be preserved also extending it with more advanced features, such as recognition of words in presence of orthographical errors. 
As a future work, we plan to provide formal and experimental comparisons of performances of \algo\ with respect to the software proposed in~\cite{Uppsala}.
}

\section{Software implementation: the user interface}\label{sec:uiandbenchmark}



\begin{figure}[t]
\centering
\includegraphics[width=\textwidth]{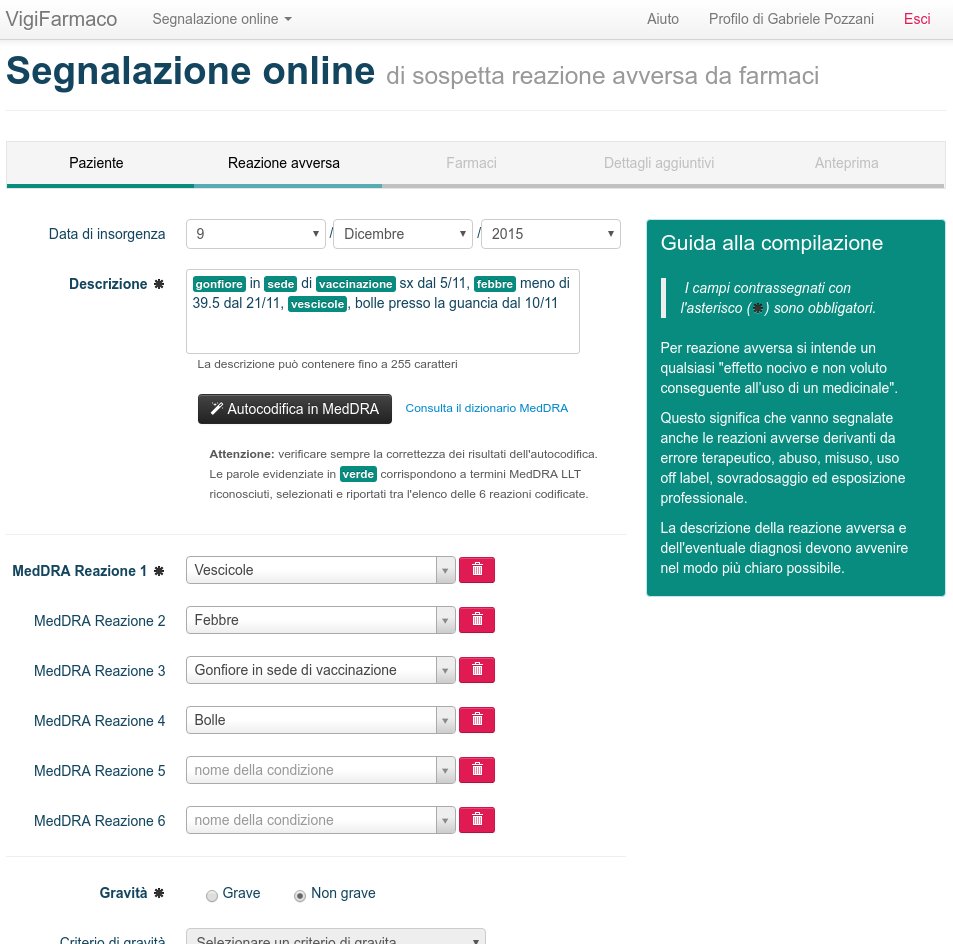}
\caption{A partial screenshot of \vigi\ User Interface}
\label{fig:vigiWScreen}
\end{figure}

\algo\ has been implemented as  a \vigi\ plug-in: people responsible for pharmacovigilance  can consider the results of the auto-encoding of the narrative description and then revise and validate it. 
Figure~\ref{fig:vigiWScreen} shows a screenshot of \vigi{} during this task. In the top part of the screen it is possible to observe the five sections composing a report. The screenshot actually shows the result of a human-\algo\ interaction: by pressing the button ``Autocodifica in \med'' (in English, ``\med\ auto-encoding''), the responsible for pharmacovigilance obtains a \med\ encoding corresponding to the natural language ADR in the field ``Descrizione" (in English, ``Description''). 
Up to six solutions are proposed as the best \med\ term candidates returned by \algo: the responsible can refuse a term (through the trash icon), change one or more terms (by an option menu), or simply validate the automatic encoding and switch to the next section ``Farmaci'' (in English,  ``Drugs''). The maximum number of six terms to be shown has been chosen in order to supply pharmacovigilance experts with a set of terms extended enough to represent the described adverse drug reaction but not so large to be redundant or excessive. 

We are testing \algo\ performances in the daily pharmacovigilance activities. Preliminary qualitative results show that \algo\ drastically reduces the amount of work required for the revision of a report, allowing the pharmacovigilance stakeholders to provide high quality data about suspected ADRs.

\section{Testing \algo\ performances}\label{sec:testing}

In this section we describe the experiments we performed to evaluate \algo\ performances. 
The  test  exploits a large amount of manually revised reports we obtained from \vigis~\cite{Vigisegn}.


We briefly recall two metrics we used to evaluate \algo: \emph{precision} and \emph{recall}. 

In statistical hypothesis and in particular in binary classification~\cite{Manning08}, two main kinds of errors are pointed out: \emph{false positive errors} (FP) and \emph{false negative errors} (FN). 
In our setting, these errors can be viewed as follows: a false positive error is the inopportune retrieval of a ``wrong'' \llt , i.e., a term which does not correctly encode the textual description; a false negative error is the failure in the recognition of a ``good'' \llt, i.e., a term which effectively encode (a part of) the narrative description and that would have been selected by a human expert. As dual notions of false positive and false negative, one can define \emph{correct} results, i.e., \emph{true positive} (TP) and \emph{true negative} (TN): in our case, a true positive is a correctly returned \llt, and a true negative is an \llt\ which, correctly, has not been recognized as a solution. 


Following the information retrieval tradition, the standard approach to  system evaluation revolves around the notion of relevant and non-relevant solution (in information retrieval, a solution is represented by a document~\cite{Manning08}). We provide here a straightforward  definition of \emph{relevant solution}. A relevant solution is a \med\ term which correctly encode the narrative description provided to \algo. 
A retrieved solution is trivially defined as an output term, independently from its relevance. We dub the sets of relevant solutions and retrieved solutions as $\mathsf{RelS}$ and $\mathsf{RetS}$, respectively.

The evaluation of the false positive and the false negative rates, and in particular of the impact of relevant solutions among the whole set of retrieved solutions, are crucial measures in order to estimate the quality of the automatic encoding.

The \emph{precision} (P), also called positive predictive value, is the percentage of retrieved solutions that are relevant. 
The \emph{recall} (R), also called sensitivity, is the percentage of all relevant solutions returned by the system. 

Table~\ref{tab:irmeasure} summarizes 
formulas for precision and recall. 
We provide formulas both in terms of relevant/retrieved solutions and false positives, true positives and false negatives.

\begin{table*}
\centering
\begin{tabular}{|c|c|}
\hline
\rule[-0.8em]{0pt}{2.2em}Precision& $\mathtt{P}=\frac{\,|\rels\;\cap\;\rets|\,}{|\rets|}=\frac{TP}{TP+FP}$ \\\hline
\rule[-0.8em]{0pt}{2.2em}Recall& $\mathtt{R}=\frac{\,|\rels\;\cap\;\rets|\,}{|\rels|}=\frac{TP}{TP+FN}$ \\\hline
\end{tabular}
\caption{Performance and correctness measures}\label{tab:irmeasure}
\end{table*}

It is worth noting that the binary classification of solutions as relevant or non-relevant is referred to as the gold standard  judgment of relevance. In our case, the gold standard has to be represented by a \emph{human encoding} of a narrative description, i.e., a set of \med\ terms choosen by a pharmacovigilance expert. Such a set is assumed to be definitively  \emph{correct}  (only correct solutions are returned) and \emph{complete} (all correct solutions have been returned). 

\subsection{Experiment about \algo\ performances}\label{sec:firstexp}
To evaluate \algo\ performances, we developed a benchmark, which automatically compares \algo\ behavior with human encoding on already manually revised  and validated \adr\ reports.

For this purpose, we exploited  \vigis, a data warehouse and OLAP system that has been developed for the Italian Pharmacovigilance National Center \cite{Vigisegn}. This system is based on the open source business intelligence suite Pentaho\footnote{http://www.pentaho.com/}.  \vigis\ offers a large number of \emph{encoded}  \adr s. The encoding has been manually performed and validated by experts working at pharmacovigilance centres. Encoding results have then been sent to the national regulatory authority, AIFA.

We performed a test composed by the following steps.
\begin{enumerate}
\item We launch an ETL procedure through Pentaho Data Integration.  Reports are transferred  from \vigis\  to an ad hoc database \textsf{TestDB}. The dataset covers all the 4445 reports  received, revised and validated during the year 2014 for the Italian region Veneto.
\item The ETL procedure  extracts  the narrative descriptions from reports stored in \textsf{TestDB}. For each description, the procedure calls \algo\ from\\ \vigi; the output, i.e., a list of \med\ terms, is stored in a table of \textsf{TestDB}. 
\item Manual and automatic encodings of each report are finally compared through an SQL query. In order to have two uniform data sets, we compared only those reports where \algo{} recognized at most six terms, i.e., the maximum number of terms that human experts are allowed to select through the \vigi{} user interface. Moreover, we map each \llt\ term recognized by both the human experts and \algo{} to its corresponding preferred term. Results are discussed below in Section~\ref{sec:test1discussion}. 
\end{enumerate}

\subsubsection{Experiment: analysis of results}\label{sec:test1discussion}

\begin{table}
\centering
{\footnotesize
    \begin{tabular}{|c | c| c |c | c | c | c | c | }
      \hline
     \rule{0pt}{1.1em}\textbf{Class}& \textbf{\# chars}&\textbf{\# reports} &\textbf{Common PT} &\textbf{FN}&\textbf{FP}&\textbf{R}&\textbf{P}\\ 
      \hline
     \rule{0pt}{1.1em}1& 0-20 chars& 459 & 86\% & 8\% & 7\% & 86\% & 88\%\\\hline
     \rule{0pt}{1.1em}2& 21-40 chars& 1012 & 68\% & 18\% & 14\% & 72\% & 75\%\\\hline
     \rule{0pt}{1.1em}3& 41-100 chars& 1993 & 51\% & 25\% & 24\% & 61\% & 62\%\\\hline
     \rule{0pt}{1.1em}4& 101-255 chars & 970 & 42\%  & 24\% & 34\% & 58\% & 52\%\\\hline
     \rule{0pt}{1.1em}5& $>$255 chars & 11 & 33\% & 32\% & 35\% & 46\% & 45\%\\\hline
    \end{tabular}
}
\caption{First results of \algo\ performances}\label{tab:percentage}
\end{table}

\begin{figure} 
\begin{center}
\includegraphics[width=\textwidth]{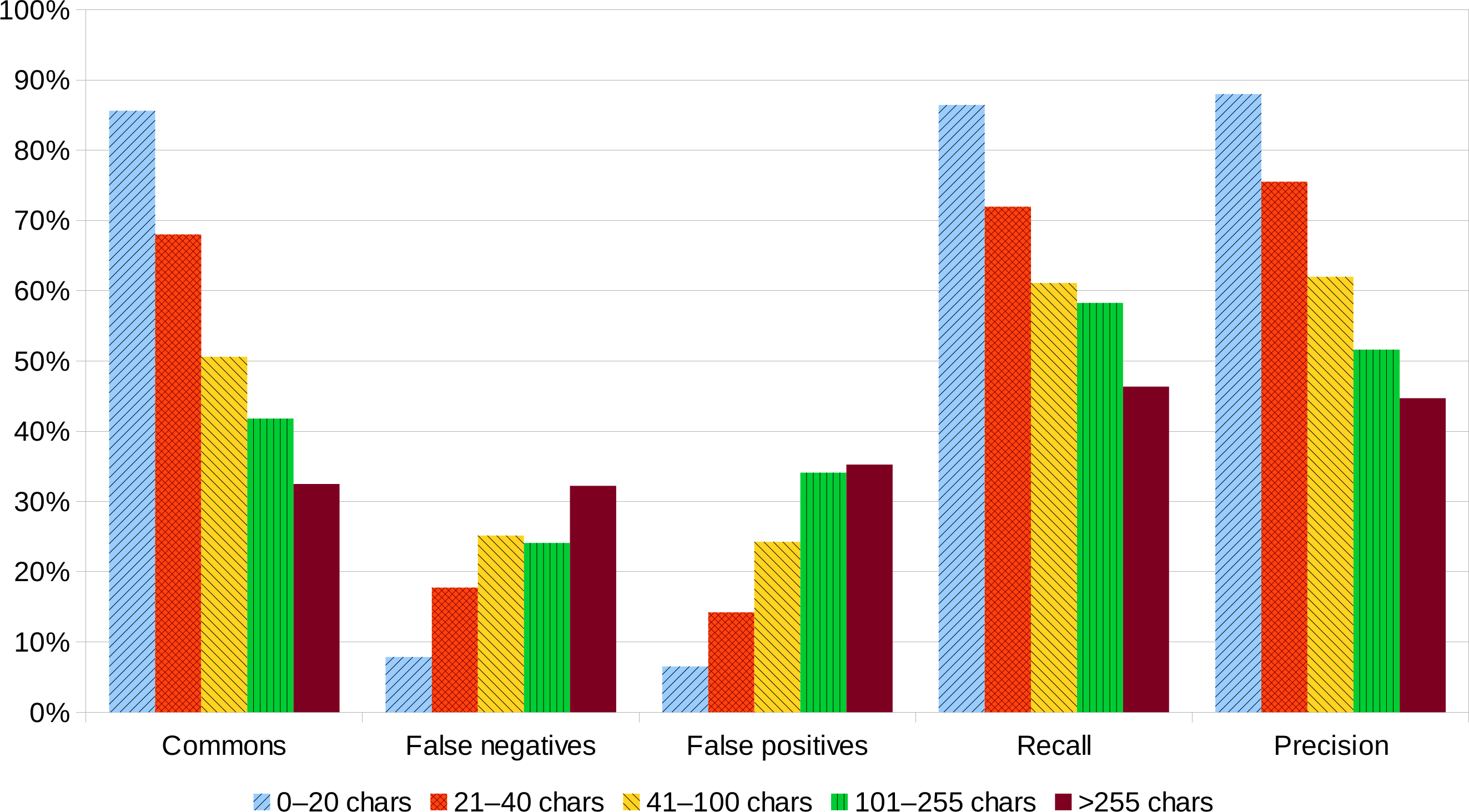}
\caption{Graphical representation of \algo\ performances}
\label{fig:performances}
\end{center}
\end{figure}

Table~\ref{tab:percentage} shows the results of this first performance test. We group narrative descriptions by increasing length (in terms of characters). We note that reported results are computed considering terms at \pt{} level. By moving to \pt{} level, instead of using the \llt{} level, we group together terms that represent the same medical concept (i.e., the same adverse reaction). In this way, we do not consider an error when \algo{} and the human expert use two different \llt{}s for representing the same adverse event. The use of the \llt{} level for reporting purpose and the \pt{} level for analysis purpose is suggested also by \med{}~\cite{MeddraGuide}. With \emph{common \pt{}} we mean the percentage of preferred terms retrieved by human reviewers that have been recognized also by \algo. 
Reported performances are summarized also in \figurename{}~\ref{fig:performances}. Note that, false positive and false negative errors are required to be as small as possible, while common \pt{}, recall, and precision have to be as large as possible.

\algo\ behaves very well on very short descriptions (class 1) and on short ones (class 2). Recall and precision remain greater than 50\% up to class 4. Notice that very long descriptions (class 5), on which performances drastically decrease, represent a negligible percentage of the whole set (less than 0.3\%). 

Some remarks are mandatory. It is worth noting that this test simply estimates how much, for each report, the \algo\ behavior is similar to the manual work, without considering the effective quality of  the manual encoding.  
Clearly, as a set of official reports, revised and sent to RNF, we assume to deal with an high-quality encoding: notwithstanding, some errors in the human encoding possibly occur. Moreover, the query we perform to compare manual and automatic encoding is, obviously, quantitative. For each \vigis\ report, the query is able to detect common retrieved terms and terms returned either by the human expert or by \algo. It is not able to fairly test \emph{redundancy} errors: human experts make some encoding choices in order to avoid repetitions. Thus, an \llt\ $t$ returned by \algo\ that has not been selected by the expert because redundant is not truly a false positive. As a significative counterpart, as previously said, we notice that some reports contain slightly human omissions/errors. This suggest the evidence that we are underestimating \algo\ performances. See the next section for some simple but significative examples.

\subsection{Examples}\label{sec:examples} 


Table~\ref{table:example} provides some examples of the behavior of \algo. We propose some free-text ADR descriptions from \textsf{TestDB} and we provide both the manual and the automatic encodings into  \llt\ terms. We also provide the English translation of the natural language texts (we actually provide a quite straightforward \emph{literal} translation).

\medskip\noindent
\begin{tabular}{p{.03\textwidth}p{.92\textwidth}}
D1: & {\tt anaphylactic shock (hypotension + cutaneous rash) 1 hour after taking the drug.}\\
D2: & {\tt swelling in vaccination location left from 11/5; temperature less than 39,5 from 11/21; vesicles, blisters around the cheek from 11/10.}\\
D3: & {\tt extended local reaction, local pain, headache, fever for two days.}
\end{tabular}



\medskip

In Table~\ref{table:example} we use the following notations: $\underline{t_1}^{n}$ and $\underline{t_2}^{n}$ are two \emph{identical} \llt{}s retrieved both by  the human and the automatic encoding; $\overline{t_1}^{n}$ and $\overline{t_2}^{n}$ are two \emph{semantically equivalent} or \emph{similar} \llt{}s (i.e., \llt{}s with the same \pt{}) retrieved by the human and the automatic encoding, respectively; we use bold type to denote terms that have been recognized by \algo\ but that have not been encoded by the reviewer; we use italic type in D1, D2, D3 to denote text recognized only by \algo. For example, in description D3, ``cefalea'' (in English, ``headache'')  is retrieved and encoded both by the human reviewer and  \algo; in description D2,  ADR ``febbre'' (in English, ``fever') has been encoded with the term itself by the algorithm, whereas the reviewer encoded it with its synonym ``piressia''; in D1,  ADR ``ipotensione'' (in English, ``hypotension'') has been retrieved only by \algo.

To exemplify how the ordered phrase heuristic works, we can notice that in D2 \algo\ did not retrieve the \med\ term ``Vescicole in sede di vaccinazione'' (10069623), Italian for ``Vaccination site vesicles''. It belongs to the set of the voted solutions (since $\cone{10069623}=0$), but it has been pruned from the list of the winning terms by the ordered-phrase heuristic criterium.

\begin{table*}
{\scriptsize\setlength\tabcolsep{3pt}\def\arraystretch{1.5}
\begin{tabular}{|p{.02\textwidth}|p{.25\textwidth}|p{.33\textwidth}|p{.33\textwidth}|}
\hline
\# & \textbf{Narrative Description} & \textbf{LLT Human Encoding}  & \textbf{LLT \algo\ Encoding}\\\hline

\hspace*{-2pt}D1 & Shock anafilattico (\textit{ipotensione} + rash cutaneo) 1 h dopo assunzione x os del farmaco&\begin{flushleft}$\underline{\mbox{Shock anafilattico}}^1$\end{flushleft} & \begin{flushleft}\textbf{Ipotensione}, $\underline{\mbox{Shock anafilattico}}^1$\end{flushleft}\\\hline

\hspace*{-2pt}D2 & gonfiore in sede di vaccinazione sx dal 5/11, febbre meno di 39,5 dal 21/11, vescicole, \textit{bolle} presso la guancia dal 10/11 & \begin{flushleft}$\overline{\mbox{Piressia}}^2$, $\underline{\mbox{Vescicole}}^3$, $\underline{\mbox{Gonfiore in sede di vaccinazione}}^1$\end{flushleft}&\begin{flushleft}\textbf{Bolle}, $\overline{\mbox{Febbre}}^2$, $\underline{\mbox{Vescicole}}^3$, $\underline{\mbox{Gonfiore in sede di vaccinazione}}^1$\end{flushleft}\\\hline

\hspace*{-2pt}D3 & Reazione locale estesa, \textit{dolore} locale; cefalea e febbre per due giorni &\vspace*{-2em}\begin{flushleft}$\underline{\mbox{Cefalea}}^1$, $\underline{\mbox{Febbre}}^2$, $\overline{\mbox{Reazione\,in\,sede\,di\,vaccinazione}}^3$\end{flushleft}\vspace*{-2em} &\vspace*{-2em}\begin{flushleft}$\underline{\mbox{Cefalea}}^1$, \textbf{Dolore}, $\underline{\mbox{Febbre}}^2$, $\overline{\mbox{Reazione locale}}^3$\end{flushleft}\vspace*{-2em}\\\hline
\end{tabular}
}
\caption{Examples of \algo\ behavior}\label{table:example}
\end{table*}

\section{Discussion}\label{sec:cons}

We discuss here some interesting points we met developing \algo. We explain the choices we made and consider some open questions.

\subsection{Stemming and performance of the NLP software}

Stemming is a useful tool for natural language processing and text searching and classification. The extraction of the stemmed form of a word is a non-trivial operation, and algorithms for stemming are very efficient. In particular, stemming for Italian language is extremely critic: this is due to the complexity of  language and the number of linguistic variations and exceptions.

For the first implementation of \algo\ as \vigi\ plug-in, we used a robust implementation of the Italian stemming procedure\footnote{http://snowball.tartarus.org/}. The procedure takes into account subtle properties of the language; in addition of the simple recognition of words up to plurals and genres, it is able, in the majority of cases, to recognize an adjectival form of a noun by extracting the same syntactical root.

Despite the efficiency of this auxiliary algorithm, we noticed that the recognition of some \med\ terms have been lost: in some sense, this stemming algorithm is too ``aggressive'' and, in some cases, counterintuitive. For example, the Italian adjective ``psichiatrico'' (in English, psichiatric) and its plural form ``psichiatrici'' have two different stems, ``psichiatr'' and ``psichiatric'', respectively. Thus, in this case the stemmer fails in recognizing the singular and plural forms of the same word.

We then decided to adopt the stemming algorithm also used in Apache Lucene\footnote{https://lucene.apache.org/}, an open source text search engine library. This procedure is less refined w.r.t. the stemming algorithm cited above, and can be considered as a ``light'' stemmer: it simply elides the final vowels of a word\footnote{A refined stemmer acts in a more complex way, taking into account also the etymological source of the words.}. This induces a conservative approach and a uniform processing of the whole set of \med\ words. This is unsatisfactory for a general  problem of text processing, but it is fruitful in our setting.
We repeated the \algo\ testing both with the classical and the light stemmer: in the latter case, we measure a global enhancement of \algo\ performance.  Regarding common retrieved preferred terms, we reveal an average enhancement of about $4\%$: percentages for classes 1, 2, 3, 4 and 5 move from $83\%$, $67\%$, $47\%$, $39\%$, $25\%$, respectively, to values in Table~\ref{tab:percentage}.
It is reasonable to think that a simple stemming algorithm maintains the recognition of words up to plurals and genres, but in most cases, the recognition up to noun or adjectival form is potentially lost. Notwithstanding, we claim that it is possible to reduce this disadvantage thanks to the embedding in the dictionary of a reasonable set of synonyms of \llt{}s (see Section~\ref{sec:syn}).

\subsection{Synonyms}\label{sec:syn} 
\algo\ performs a pure syntactical recognition (up to stemming) of words in the narrative description: no semantical information is used in the current version of the algorithm. In written informal language, synonyms are frequently used. 
A natural evolution of our NLP software may be the addition of an Italian thesaurus dictionary. 
This would appear a trivial extension: 
one could try to  match \med\  both with original words and their synonyms, and try to maximize the set of retrieved terms. We performed a preliminary test, and we observed a drastic deterioration of \algo\ performances (both in terms of correctness and completeness): on average, common PT percentages decreases of 24\%. 
The main reason is related to the nature of Italian language: synonymical groups include words related by figurative meaning. For example, among the synonyms of the word ``faccia'' (in English, ``face''), one finds  ``viso'' (in English ``visage''), which is semantically related,  but also ``espressione'' (in English, ``expression''), which is not relevant in the considered medical context. Moreover, the use of synonyms of words in ADR text leads to an uncontrolled growth of the voted terms, that barely can be later dropped in the final terms release. Furthermore, the word-by-word recognition performed by \algo, with the uncontrolled increase of the processed tokens (original words plus synonyms plus possible combinations), could induce a serious worsening of the computational complexity. Thus, we claim that this is not the most suitable way to address the problem and the designing of an efficient strategy to solve this problem is not trivial. 

We are developing a different solution, working side-by-side with  the pharmacovigilance experts. The idea, vaguely inspired by the Consumer Health Vocabulary (recalled in Section~\ref{sec:related} and used in~\cite{Yang12}),  is to collect a set of \emph{pseudo}-\llt{}s, in order to enlarge the  \med\ official terminology and to generate a new ADR lexicon. This will be done on the basis of frequently retrieved locutions which are semantically equivalent to \llt s. A pseudo \llt\ will be regularly voted and sorted by \algo\ and, if selected, the software will release the official (semantically equivalent) \med\ term. 
Notice that, conversely to the single word synonyms solution, each pseudo-\llt\ is related to one and only one official term: this clearly controls the complexity deterioration. 
Up to now, we added to the official \med\ terminology a set of about 1300 locutions. We automatically generated such a lexicon by considering three nouns that frequently occur in \med, ``aumento'', ``diminuzione'' e ``riduzione'' (in English ``increase'', ``decrease'', and ``reduction'', respectively) and their adjectival form. For each \llt\ containing one of these nouns (resp., adjectives) we generate an equivalent term taking into account the corresponding adjective (resp., noun).

This small set of synonyms induces a global improvement of \algo\ performances on classes 4 and 5. For Class 4, both common retrieved \pt{} percentage, precision and recall increase of $1\%$. For Class 5, we observe some significative increment: common retrieved \pt{} moves from $33\%$ to $37\%$; precision moves from $45\%$ to $49\%$; recall moves from $46\%$ to $55\%$.

Also false negative and false positive rates suggest that the building of the \med-thesaurus is a promising extension. False negatives   move  from $23\%$ to $22\%$ for Class 4 and  from $32\%$ to $29\%$ for Class 5. False positive percentage decrease of $1\%$ both for Class 4 and Class 5.

Class 5, which enjoys a particular advantage from the introduction of the pseudo-\llt{}s, represents a small slice of the set of reports. Notwithstanding, these cases are very arduous to address, and we have, at least, a good evidence of the validity of our approach.

\subsection{Connectives in the narrative descriptions}\label{sec:connectives}

As previously said, in \algo\ we do not take into account the \emph{structure} of written sentences. In this sense, our procedure is radically different from those based on the so called part-of-speech (PoS)~\cite{Felice09},  powerful methodologies able to perform the morpho-syntactical  analysis of texts, labeling each lexical item with its grammatical properties.
PoS-based text analyzers are also able to detect and deal with logical connectives such as conjunctions, disjunctions and negations. Even if connectives  generally play a central role in the logical foundation of natural languages, they have a minor relevance in the problem we are addressing: ADR reports are on average badly/hurriedly written, or they do not have a complex structure (we empirically noted this  also for long descriptions).
Notwithstanding, negation deserves a distinct consideration, since the presence of a negation can drastically change the meaning of a phrase. 
First, we evaluated the frequency of negation connectives in ADR reports:  we considered the same sample exploited in Section~\ref{sec:firstexp}, and we counted the occurrences of the words ``non'' (Italian for ``not'') and ``senza'' (Italian for ``without'')\footnote{The word ``senza'' does not necessarily imply a negation, thus we are clearly overestimating the presence of negations.}: we detected potential negations  in 162 reports (i.e., only in the $3.5\%$ of the total number, 4445). 
Even though negative sentences seem to be uncommon in ADR descriptions, the detection of negative forms is  a short-term issue we plan  to address. 
As a first step, we plan to recognize words that may represent negations and to signal them to the reviewer through the graphical UI. In this way, the software sends to the report reviewer an alert about the (possible) failure of the syntactical recognition.

\subsection{On the selection of voted terms} 

As previously said, in order to provide an effective support to human revision work, it is necessary to provide only a small set of possible solutions. To this end, in the selection phase (described in Section~\ref{sec:winning}), we performed drastic cuts on voted \llt\mbox{}s. For example, only completely covered \llt\mbox{}s can contribute to the set of winning terms. This is clearly a restrictive threshold, that makes completely sense in a context where at most six solutions can be returned. In a less restrictive setting, one can relax the threshold above and try to understand how to filter more ``promising'' solutions among partially covered terms. In this perspective, we developed a further criterion, the \emph{Coverage Distribution}, based on assumptions we made about the structure of  (Italian) sentences. 
The following formula simply sums the indexes of the covered words for  $t\in\vllt$:

\begin{displaymath}
\cfive{t}=\sum_{i=0}^{\voted{t}.length-1} \voted{t}[i]
\end{displaymath}

If $\cfive{t}$ is small, it means that words in the first positions of term $t$ have been covered. We defined $\cfive{\cdot}$ to discriminate between possibly joint winning terms. Indeed, an Italian medical description of a pathology has frequently the following shape: \emph{name of the pathology}+\emph{``location'' or adjective}. Intuitively, we privilege terms for which the recognized words are probably the ones describing the pathology.
The addition of $\cfive{\cdot}$ (with the discard of condition $\cone{\dot}=0$ in the final selection) could improve the quality of the solution if a larger set of winning terms is admissible or in the case in which the complete ordered list of voted terms is returned.

\section{Conclusions and future work}\label{sec:future}

In this paper we proposed \algo, a simple and efficient NLP software, able to provide a concrete support to the pharmacovigilance task, in the revision of ADR spontaneous reports. \algo\ takes in input a narrative description of a suspected ADR and produces as outcome a list of \med\ terms that ``covers'' the medical meaning of the free-text description. Differently from other BioNLP software proposed in literature, we developed an original text processing procedure. 
Preliminary results about \algo\ performances are encouraging. 
Let us sketch here some ongoing and future work. 

We are addressing the task to include ad hoc knowledges, as the \med-thesaurus described in Section~\ref{sec:syn}. We are also proving that \algo\ is robust with respect to language (and dictionary) changes.  The way the algorithm has been developed suggests that \algo\ can be a valid tool also for narrative descriptions written in English. Indeed,  the algorithm retrieves a set of words, which covers an \llt\ $t$, from a free-text description, only slightly considering the order between words or the structure of the sentence. This way, we avoid the problem of ``specializing'' \algo\ for any given language. 
We plan to test \algo\ on the English \med\ and, moreover, we aim to test our procedure on different dictionaries (e.g., ICD-9 classification\footnote{http://icd9cm.chrisendres.com/}, WHO-ART\footnote{https://www.umc-products.com}, SNOMED CT\footnote{http://www.ihtsdo.org/snomed-ct/what-is-snomed-ct}). 
We are collecting several sources of manually annotated corpora, as potential testing platforms. 
Moreover, we plan to address the  management of orthographical errors possibly contained in narrative ADR descriptions. We did not take into account this issue in the current version of \algo. A solution could include an ad hoc (medical term-oriented) spell checker in \vigi, to point out to the user that she/he is doing some error in writing the current word in the free description field. This should drastically reduce users' orthographical errors without  
{heavy side effects} in \algo\ development and performances.
Finally, we aim to apply  \algo\ (and its refinements) to different sources for ADR detection, such as drug information leaflets and social media~\cite{Yang12,Sarker2015}.

\end{document}